\pdfoutput=1
\documentclass[11pt,a4paper]{article}
\usepackage{emnlp2021}
\usepackage{times}
\usepackage{latexsym}

\usepackage[T1]{fontenc}
\usepackage[utf8]{inputenc}

\usepackage{microtype}
\usepackage{amsmath}
\usepackage{hyperref}
\usepackage{graphicx,amsmath,amssymb}
\usepackage{booktabs}
\usepackage{subcaption}
\usepackage{tikz-qtree}
\usepackage{enumitem}
\usetikzlibrary{positioning}
\usepackage{bm}
\usepackage{multirow}
\usepackage{bbm}
\usepackage{textcomp}
\usepackage{pgfplots}
\usepackage{tabularx}
\usepackage{algorithm} 
\usepackage{algpseudocode} 
\algnewcommand{\Inputs}[1]{%
  \State \textbf{Inputs:}
  \Statex \hspace*{\algorithmicindent}\parbox[t]{.8\linewidth}{\raggedright #1}
}
\algnewcommand{\Initialize}[1]{%
  \State \textbf{Initialize:}
  \Statex \hspace*{\algorithmicindent}\parbox[t]{.8\linewidth}{\raggedright #1}
}
\pgfplotsset{width=8cm, compat=1.16}

\definecolor{DeepGreen}{RGB}{62, 67, 195}
\definecolor{DeepBlue}{RGB}{57, 115, 77}
\definecolor{DeepRed}{RGB}{141, 59, 67}
\definecolor{Revision}{RGB}{95, 26, 55}

\newcommand{\ours}{\textsc{Concrete L-Pcfg}}
\newcommand{\coupling}{\textsc{Coupling}}
\newcommand{\LPCFG}{neural L-PCFG}

\newcommand{\thicklegend}{\raisebox{2pt}{\tikz{\draw[->,black, solid, line width = 1.5pt](0,0) -- (5mm,0);}}}

\newcommand{\dottedlegend}{\raisebox{2pt}{\tikz{\draw[->,black, dotted, line width = 1.5pt](0,0) -- (5mm,0);}}}

\title{Dependency Induction Through the Lens of Visual Perception}

\author{Ruisi Su$^1$\hspace{2em}
Shruti Rijhwani$^2$\hspace{2em}
Hao Zhu$^2$\hspace{2em}
Junxian He$^2$ \\
\textbf{Xinyu Wang$^2$}\hspace{2em}
\textbf{Yonatan Bisk$^2$} \hspace{2em} \textbf{Graham Neubig$^2$} \\
$^1$Electrical and Computer Engineering, Carnegie Mellon University \\
$^2$Language Technologies Institute, Carnegie Mellon University \\
{\tt ruisis@alumni.cmu.edu, zhuhao@cmu.edu} 
}

\date{}

\begin{document}
\maketitle
\begin{abstract}

Most previous work on grammar induction focuses on learning phrasal or dependency structure purely from text. However, because the signal provided by text alone is limited, recently introduced visually grounded syntax models make use of multimodal information leading to improved performance in constituency grammar induction.
However, as compared to dependency grammars, constituency grammars do not provide a straightforward way to incorporate visual information without enforcing language-specific heuristics.
In this paper, we propose an unsupervised grammar induction model that leverages word concreteness and a structural vision-based heuristic to jointly learn constituency-structure and dependency-structure grammars. 
Our experiments find that concreteness is a strong indicator for learning dependency grammars, improving the direct attachment score (DAS) by over 50\% as compared to state-of-the-art models trained on pure text. 
Next, we propose an extension of our model that leverages both word concreteness and visual semantic role labels in constituency and dependency parsing. Our experiments show that the proposed extension outperforms the current state-of-the-art visually grounded models in constituency parsing even with a smaller grammar size.\footnote{Code is available at \url{https://github.com/ruisi-su/concrete_dep}.}
\end{abstract}

\section{Introduction}
\label{sec:intro}

Grammar induction aims to discover the underlying grammatical structure of a language from strings of symbols. Previous work has focused on two grammar formalisms: constituency and dependency grammars. 
Probabilistic context-free grammars \cite[PCFG]{10.5555/1864519.1864540, clark-2001-unsupervised} have been used for modeling probabilistic rules in a constituency grammar, including more recent approaches that combine PCFGs with neural models~\cite{kim-etal-2019-compound}. The dependency grammar approach to syntactic modeling allows for a binary probabilistic treatment between each word and its head. 
This binary relationship introduces flexibility for modeling languages with rich morphology and relatively free word order languages~\cite{Jurafsky:2009:SLP:1214993}. 
A commonly used model for dependency induction is the dependency model with valence \cite[DMV]{ klein-manning-2004-corpus}.

\tikzstyle{lpcfg}=[draw=black,thick,anchor=west]
\tikzstyle{concr}=[dashed]
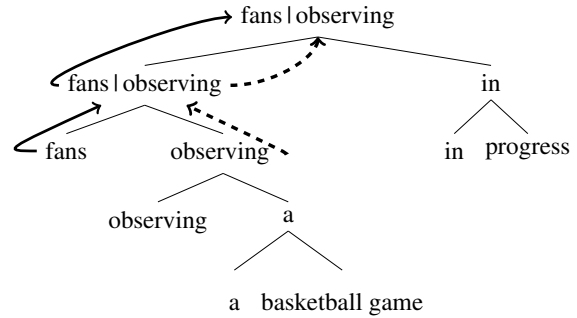
\begin{figure}[t]
    \centering
    \resizebox{\linewidth}{!}{%
    \begin{tikzpicture}
    \tikzset{every tree node/.style={align=center,anchor=north}}
        \Tree [.\node(root){fans\,|\,observing}; [.\node(np){fans\,|\,observing}; [.\node(sub){fans}; ]
            [.\node(vp){observing }; observing
               [.a \\a \\{basketball game} ] ] ]
               [.in [.\node(in){in};  ] [.\node(prog){progress};  ] ] 
                ]
    \draw[dashed,->,line width = 1.5pt] (vp)..controls +(east:1) ..(np);
    \draw[dashed,->,line width = 1.5pt] (np)..controls +(east:2) and + (south:0.6) ..(root);
    \draw[very thick,->] (sub)..controls +(west:1) ..(np);
    \draw[very thick,->] (np)..controls +(west:2) and + (west:2) ..(root);
    \end{tikzpicture}  
    }
    \caption{A learned tree structure with its lexicalized heads. Solid lines represent the constituents. Arrows represent the propagation path of the lexical heads from the constituents to the root.
    \dottedlegend \, and \thicklegend \, are the paths of the \LPCFG{}~which uses pure text and our proposed \ours{} which uses word concreteness, respectively. \LPCFG{} incorrectly predicts \textit{observing} as the root of the sentence, while \ours{} selects the correct root \textit{fans}.
    }
    \vspace{-2mm}
    \label{fig:treestruct}
\end{figure}

The two formalisms are complementary and unified models have been shown to achieve higher induction accuracy than models that are trained for either constituency or dependency parsing alone. For instance, the lexicalized PCFG \cite{collins-2003-head} is a supervised model that jointly learns the two formalisms by associating a word and a part-of-speech tag with each non-terminal in the tree of the PCFG. A more recent variant, the neural L-PCFG~\cite{zhu-etal-2020-return} is an unsupervised model that uses lexical information to jointly learn both constituency-structure and dependency-structure grammars, achieving state-of-the-art performance in both formalisms.

Besides the use of lexical information, grammar induction also benefits from a coarse understanding of the similarities in the syntactic structures shared by many human languages. Recent works showed that biases based on such syntactic universals 
can be applied to the posterior distribution of the dependency structures~\cite{naseem-etal-2010-using,cai-etal-2017-crf, Li_Cheng_Liu_Keller_2019}. Such universal linguistic rules can be helpful when part-of-speech tags are given to the grammar induction models. However, explicitly providing a ruleset can make non-terminals sensitive to their context, which is slightly contradictory to context-free grammar based methods, such as PCFGs. Thus, a more abstract method is needed to provide both guidance and flexibility to grammar induction models.

Nevertheless, inducing grammars entirely from pure text is a tall order, and arguably one not even tackled by humans themselves -- humans learn language from the surrounding world \cite{bisk2020experience}. Recent work, such as the Visually Grounded Neural Syntax Learner \cite[VGNSL]{shi2019visually} show improved performance on constituency grammar induction by learning a shared embedding space for text and its corresponding image.
However, \citet{kojima-etal-2020-learned} note that the improvement in performance is not a result of learning complex syntactic rules, and is rather of the model indirectly learning word \emph{concreteness}, a concept that evaluates the degree to which a word refers to a perceptible entity \cite{concrete}. 

 \begin{figure}
    \includegraphics[trim=0 30 0 10,clip,width=\linewidth]{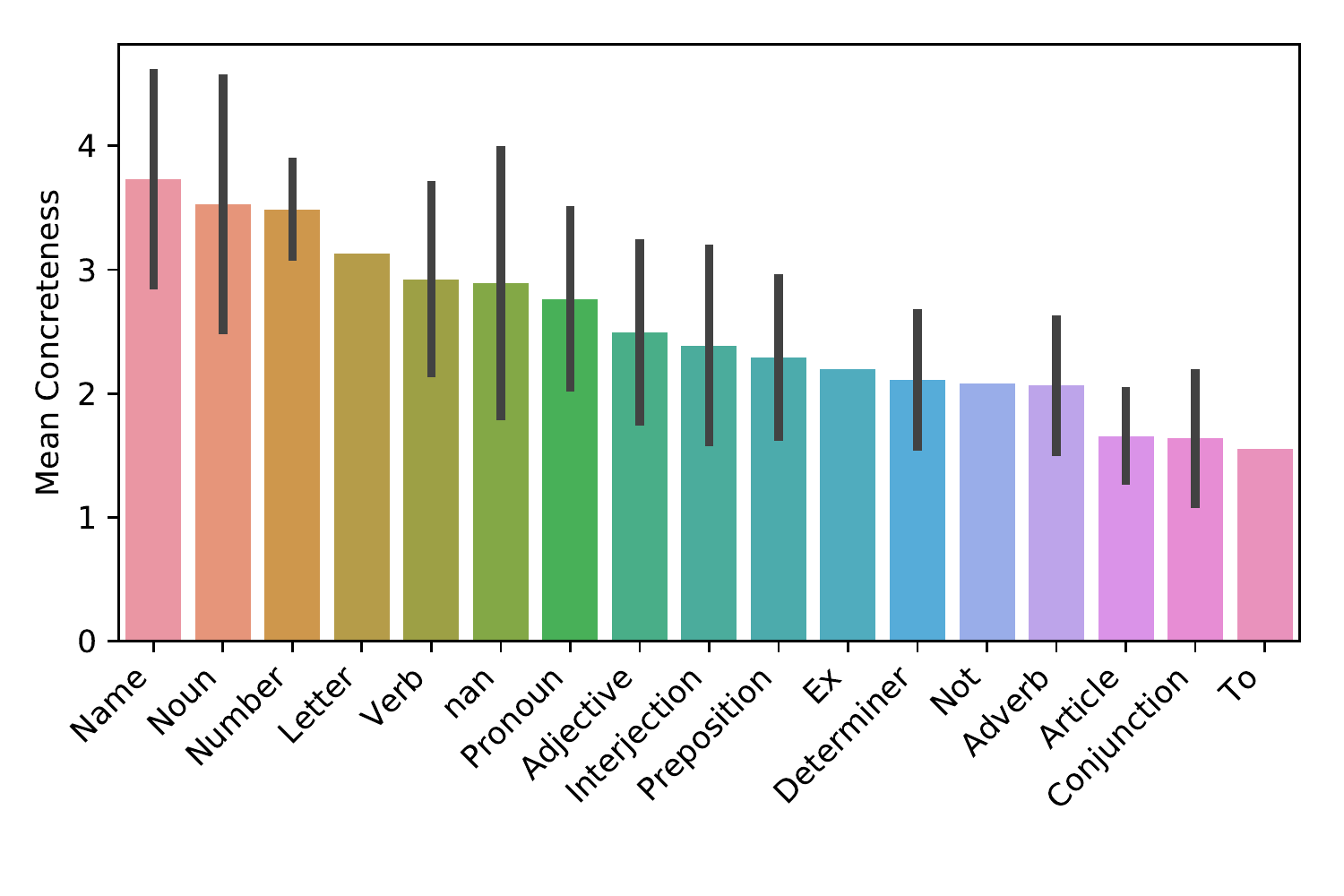}
    \caption{Mean word concreteness by dominant part-of-speech tags~\cite{concrete} with standard deviations of the English MSCOCO captioning data~\cite{DBLP:journals/corr/LinMBHPRDZ14}. We exclude words that are unclassified or lack a dominant tag.}
\label{fig:sort}
\end{figure}

However, concisely describing the effect of word concreteness on the constituency structure of a sentence is non-trivial without resorting to language-specific heuristics such as the head-initial bias of the VGNSL~\cite{shi2019visually}.
On the other hand, describing the effect of concreteness on \emph{dependency} structure is more straightforward.
In most languages, content words, such as nouns and verbs, tend to be more \textit{concrete} as they are learned through experience via modalities~\cite{concrete}. Our analysis on the English MSCOCO dataset, shown in \autoref{fig:sort}, affirms this is true for English.
Content words are also more likely to be heads than dependents in many dependency specifications~\cite{nivre-etal-2016-universal}.

Motivated by these observations, in this paper, we incorporate the concept of concreteness in the unsupervised neural L-PCFG model~\cite{zhu-etal-2020-return} at two levels: the word level and the phrase level. First, at the word level, we incorporate a prior to upweight dependency structures that have heads with higher word concreteness scores, which are human-rated numerical values for common English words~\cite{concrete}. Our experiments on the English MSCOCO dataset show that, at the word level, the word concreteness priors greatly improve the dependency induction performance over the neural L-PCFG, increasing the directed attachment accuracy (DAS) by 50\%~(\autoref{sec:exp}). Additionally, we investigate the effect on predicted roots and find that the concreteness priors serve as a ``short-cut'' to bias perceptible entities towards becoming heads in the dependency parse. In the example in \autoref{fig:treestruct}, the \ours{} selects the entity \textit{fans} as the root word for the sentence, because it is more perceptible than the action \textit{observing}. Adding word concreteness priors to the root encourages the subsequently learned rules to select the lexical head with higher word concreteness. 

At the phrase level, we present a vision-based heuristic to exploit concreteness by connecting the word referents of the corresponding perceptible entities. These perceptible entities are extracted from images and aligned using a unified vision-language pre-training model~\cite[VLP]{Zhou_Palangi_Zhang_Hu_Corso_Gao_2020}. Our experiments show the vision-based heuristic improves the model's constituency parsing performance. 
Finally, we present a model that combines the concreteness priors at both the word and phrase levels, which further improves the F1 score by 12\% over the neural L-PCFG for the constituency parsing task. 

In summary, this paper demonstrates that concreteness interacts with the neural L-PCFG differently at various levels: word-level concreteness helps dependency induction in differentiating between content and function words, and the phrase-level vision-based heuristic helps pruning grammars that produce incorrect spans. Both levels help the neural L-PCFG to conceptualize the relationships between words that are highly associated with visually perceptible entities. 

\section{Background}
\label{sec:bg}

This section briefly discusses the neural L-PCFG~\cite{zhu-etal-2020-return}, which jointly learns constituency-structure and dependency-structure grammars. We encourage readers to refer to the original paper for a more detailed description.


The neural L-PCFG is a neural parameterization of the lexicalized PCFG~\cite[L-PCFG]{collins-2003-head} and it demonstrates improved performance on unsupervised constituency and dependency parsing. The phrase structures from a context-free grammar (CFG) are defined as a five-tuple
$\mathcal{T} = \left(S,\mathcal{N}, \mathcal{P}, \Sigma, \mathcal{R}\right)$, including the following set of rules $\mathcal{R}$:
\begin{alignat*}{3}
    S &\rightarrow A \left[\alpha\right], && A \in \mathcal{N} 
    \\
    A\left[\alpha\right] & \rightarrow B\left[\alpha\right] C\left[\beta\right] , \quad && A \in \mathcal{N},  B, C \in \mathcal{N} \cup \mathcal{P}
    \\
    A\left[\alpha\right] &\rightarrow B\left[\beta\right] C\left[\alpha\right] , && A \in \mathcal{N}, B, C \in \mathcal{N} \cup \mathcal{P}
    \\
    T\left[\alpha\right] & \rightarrow \alpha , && T \in \mathcal{P}
\end{alignat*}
where $\mathcal{N}$ and $\mathcal{P}$ represent the set of non-terminals and preterminals, respectively, and $\alpha,\beta \in \Sigma$ where $\Sigma$ is the set of lexical items in grammar $\mathcal{T}$. The branching rules' scores rely on four components: (1) the probability of the root to the non-terminal $A$, (2) the word emission probability, (3) the probability of the headedness direction and the head-inheriting child conditioned on the parent non-terminal and head words, and (4) the probability of the non-inheriting child conditioned on the headedness direction, parent non-terminal, and head-inheriting child non-terminals. 

From \citet{zhu-etal-2020-return}, the neural L-PCFG 
is a parameter-sharing method that allows more flexible parameterization than traditional L-PCFGs by conditioning the probabilities of production rules on the representations of non-terminals, preterminals, and lexical items. The neural L-PCFG also utilizes the compound probability distribution stemming from the compound PCFG \cite{kim-etal-2019-compound}, which showed promising results in estimating the parameters of the model based on natural linguistic differences in the inputting sentences. The latent compound variable $\bm{z}$ is sampled from a standard spherical Gaussian distribution, whose probability is denoted as $p_{\mathcal{N}(\bm{0}, \bm{I})}(\bm{z})$. The log likelihood of a sentence $\bm{x}$ is obtained by marginalizing the compound variable:

\begin{equation}
    \label{eq:px}
    \log p(\bm{x}) = \log \int_{\bm{z}} p_{\bm{z}}(\bm{x})p_{\mathcal{N}(\bm{0}, \bm{I})}(\bm{z}) \,dz
\end{equation}

Because it is intractable to integrate over $\bm{z}$ in \autoref{eq:px}, the evidence lower bound (ELBo) of the log likelihood is estimated by Monte-Carlo sampling and optimized during training of the neural L-PCFG \cite{zhu-etal-2020-return} using the mean and the variance vectors predicted by an inference network. The neural L-PCFG is trained to maximize the log likelihood of the entire corpus.

During inference, the most probable tree is approximated using the mean vector $\bm{\mu} = f_{\mu}(\bm{x})$ (predicted by the inference network) in a Dirac delta distribution $\delta(\bm{z} - \bm{\mu})$ instead of the real distribution $p(\bm{z} | \bm{x})$, which would be intractable: 
\begin{align*}\label{eq:approx}
    \hat{t} &\approx \arg\max_{t \in \mathcal{T}_{\bm{x}}} \int_{\bm{z}} p_{\bm{z}}\delta(\bm{z} - \bm{\mu})\,dz \\
     &= \arg\max_{t \in \mathcal{T}_{\bm{x}}} p_{\bm{\mu}}(t)
\end{align*}

The probability distribution over sentences $\bm{x}$ is then defined by:
\begin{align*}
    p_{\bm{z}}(\bm{x}) = \sum_{t \in \mathcal{T}_{\bm{x}}}
    p_{\bm{z}}(t) = \frac{1}{\mathcal{Z}(\mathcal{T},
    \bm{z})} \sum_{t \in \mathcal{T}_{\bm{x}}} \tilde{p}_{\bm{z}}(t)
\end{align*}
in which $\mathcal{Z}(\mathcal{T}, \bm{z})$ is the normalizing factor, and  $\sum_{t \in \mathcal{T}_{\bm{x}}}p_{\bm{z}}(t)$ is the sum of normalized probabilities of trees that might have generated $\bm{x}$. The unnormalized probability is calculated by exponentiating a scoring function: $\tilde{p}_{\bm{z}}(t) \propto \exp G_{\theta} (t, \bm{z})$, which assumes each rule is independent of the other rules in the parse tree $t$. 

\section{Dependency Induction Method}
\label{sec:model}
We propose three model variants: the word concreteness model, the vision-based heuristic model, and a model that combines both concreteness and the heuristic for inducing dependency structures and constituency structures. 

\begin{figure*}
\noindent
  \begin{minipage}[b]{0.4\textwidth}
    \centering
    \includegraphics[width=0.9\textwidth]{figures/COCO_val2014_000000223094.jpg}
    \subcaption{The example image
    \label{fig:subalign}
    }
  \end{minipage}
    \hspace*{14pt}%
    \noindent
  \begin{minipage}[b]{0.52\textwidth}
        \begin{small}
        \noindent\begin{tabularx}{\textwidth}{@{}lr@{}}
        \toprule
          \multicolumn{2}{c}{source: (a) kitchen (with) (a) stove oven (and) refrigerator} \\ \midrule
          \multicolumn{2}{c}{target: overflowing refrigerator kitchen water} \\ \midrule 
          alignment pair  & scores \\ \midrule
            kitchen:kitchen & 0.625 \\
                                 refrigerator:refrigerator & 0.363 \\
                                 stove:overflowing & 0.030 \\
                                 oven:water & 0.006 \\
            \bottomrule 
        \end{tabularx}%
        \subcaption{The example input and output of the alignment method. }
        \label{tab:subalign}
        \end{small}
  \end{minipage}
  \caption{An example of an image-caption pair from the MSCOCO test split. \autoref{tab:subalign} shows the source (captions) and target (semantic role labels) for the alignment method of \autoref{fig:subalign}. Words in parentheses are the stop words removed from the original caption. We use the SpaCy default list of stop words for English~\cite{spacy2}. The labels are ordered by the first term as the predicted activity followed by the set of the entities involved in the activity~\cite{yatskar2016}. The alignment pairs are ordered from the highest to the lowest scores.}
  \label{fig:align}
\end{figure*}
\subsection{Proposed Model}
We extend the neural L-PCFG model to incorporate the concept of concreteness by adding two concreteness priors to the context-free scoring function:
\begin{equation}\label{eq:main_eq}
     G_{\theta}(t, \bm{z}) = \sum_{i=1}^k g_{\theta}(r_{i}, \bm{z}) 
     + \lambda_c h_{i} 
     + \lambda_{v} \mathbf{1_{v}}(s_i)
\end{equation}
For a parsed tree $t$, $g_{\theta}(r_{i}, \bm{z})$ is the original scoring function of the neural L-PCFG (\autoref{sec:bg}) that assigns a log-likelihood to rule $r_{i}$.
$\lambda_c$ is the hyperparameter used to control the effect of \textbf{the word-level concreteness prior} $h_{i}$ of the current head word $i$, detailed in \autoref{sec:concretemodel}.
Similarly, $\lambda_v$ is the hyperparameter controlling the effort of \textbf{the phrase-level concreteness prior}. $\mathbf{1_{v}}(s_i)$ is the indicator function $\left[s_i \in \mathcal{V}\right]$ that the parsed constituent $s_i$ is in the set of the rewarded spans $\mathcal{V}$ generated using the vision-based coupling heuristic in~\autoref{sec:couplingmodel}.

\subsection{Concrete L-PCFG}
\label{sec:concretemodel}
The first variant of the model we experiment with is denoted by \ours{}. In this variant, we set $\lambda_v=0$ in \autoref{eq:main_eq}, thus \textit{only} incorporating the word concreteness prior into the neural L-PCFG. 
The \ours{} uses the concreteness score derived from~\citet{concrete}, normalized (the raw scores are on a 5-point rating scale) to lie between 0 and 1. Given an observed sentence $\bm{x}$, $h_i$ adds the concreteness score of the root word ${x_j}$, specifically the head of the constituent generated from the rule that spans from the beginning to the end of the sentence. \autoref{tab:numbers} shows the concreteness scores of the possible lexical heads in \autoref{eq:main_eq} on an example sentence.  
\begin{table}
\centering
\resizebox{\columnwidth}{!}{
\begin{small}
\begin{tabular}{l@{\ \ \ \ }c@{\ \ }c@{\ \ }c@{\ \ }c@{\ \ }c@{\ \ }c@{\ \ }c@{}}
fans   & observe & a     & basketball & game & in  & progress \\
\midrule
0.942  & 0.526    & 0.292 & 0.994       & 0.9  & 0.6   & 0.424  \\
\end{tabular}
\end{small}
}
\caption{The concreteness score for each word in an example phrase. Perceptible entities like \textit{fans} and \textit{game} have higher scores than \textit{in} and \textit{observe}. Note that the concreteness scores are not marginalized by the words in a sentence.}
\label{tab:numbers}
\vspace{-1em}
\end{table}

\subsection{Vision-based Coupling Heuristic}

\label{sec:couplingmodel}

For the second variant of our model, we introduce the \coupling{} heuristic: instead of upweighting the root word with high word concreteness, we incentivize rules that might have generated spans according to the argument-predicate relationship extracted from the visual information. Specifically, this objective aims to reward spans that contain both an argument and predicate in a relation with the head assigned to the predicate. For example, given a caption, \textit{a girl is eating a slice of cheesecake}, and the predicted semantic role labels from the corresponding image are the agent \textit{woman}, the activity \textit{eat}, and the item \textit{cake}. \coupling{} sets $\lambda_c=0$ in \autoref{eq:main_eq}. We use a VLP \cite{Zhou_Palangi_Zhang_Hu_Corso_Gao_2020} fine-tuned on a subset of MSCOCO and a subset of imSitu~\cite{yatskar2016} to generate semantic role labels for the image corresponding to the text caption (see \autoref{fig:align}). In \autoref{eq:main_eq}, we use $s_i$ to denote the span generated by the rule $r_i$, and $\mathcal{V}$ to denote the set of the rewarded spans. The rewarded spans are selected following the procedure described below.

\paragraph{Label-Caption Alignment} For a given image, its semantic role labels consist of an action, its participants (such as actors, objects, substances, locations, and etc.), and the roles these participants play in the activity~\cite{yatskar2016}. We perform alignment between the captions of the dataset and the predicted semantic labels to locate the predicted perceived entities from the VLP model in the caption. Since the VLP model is pretrained with an unsupervised learning objective, using the vision-based heuristic on the neural L-PCFG remains an unsupervised task. The semantic labels are preprocessed before the alignment by removing the role labels. The first entity is always the main activity followed by the participants in the order they appear in the semantic labeling predictions. We also removed stop words from the captions to generate more reliable alignment pairs. This is largely because a misalignment, such as aligning determiners with nouns, can potentially be detrimental to use the \coupling~heuristic on the neural L-PCFG. An example of the alignment inputs and the generated outputs with the corresponding alignment scores are described in \autoref{fig:align}. We use the Dice alignment method~\cite{melamed1997wordtoword} to find co-occurrences of the caption words and semantic labels and generate alignment pairs between them. 

Because semantic role labels explain the situation depicted in the image, it is reasonable to couple the words of the corresponding caption that realizes the role labels in a similar way: couple the predicate representative, \textit{eat}, with the representatives of its arguments, \textit{girl} and \textit{cheesecake}. 

\begin{algorithm}[t]
\small
	\caption{Rewarded Span Generation} 
	    \label{coupling:pseudo}
	\begin{algorithmic}
	    \Inputs{$caption, labels, aligns$}
	    \Initialize{$\mathcal{V} \gets \varnothing $, $F = zeros(labels)$}
		\For { $l_1$ in $labels$}
			\For {$a$ in $aligns$}
				\State Get $cap$, $l_2$, $score$ from $a$
				\If {$l_1 == l_2$}
				\State $F_i = index(cap)$ 
				\EndIf
			\EndFor
		\EndFor
		\State compute distance $D$ between $F_0$ and rest of $F$
		\State $F_0$ is the predicate
		\State $A \gets argsort(D)$
		\For{ $i = 1,\dots,N$}
		    \State $d \gets D[A[i]]$
    		\If{$d < 0$}
    		\State $s \gets (F[A[i]], F[0], F[0])$
    		\Else{$s \gets (F[0], F[A[i]], F[0])$}
    		\EndIf
    		\State $\mathcal{V} \gets \mathcal{V} + s $
		\EndFor
	\end{algorithmic} 
\end{algorithm}

\paragraph{Rewarded Span Generation} For each caption, we use the alignment pairs to generate a set of rewarded spans $\mathcal{V}$ in the third term of \autoref{eq:main_eq}. Each generated span couples only one argument with the predicate: the start and end of the span are based on the order in which the predicate and the alignment appeared in the caption, and the predicate word is always the head word of the span. \autoref{coupling:pseudo} details the procedure for generating the rewarded spans.





\section{Experiments}
\label{sec:exp}
We run experiments for the three model variants we propose in the previous section: the word concreteness prior, the vision-based heuristic, and the combined method.
For word concreteness, our experiments aim to answer the following questions: (1) ``does incorporating concreteness in the learning process improve dependency induction?'' and (2) ``once concreteness has been used in learning, can the model still achieve good performance when concreteness information is no longer available during inference?''. In particular, the second question asks whether the model can internalize the concept of concreteness during training and accurately induce dependency structure even when concreteness is not explicitly provided. 
Additionally, we experiment with the vision-based heuristic model to explore (3) whether structured priors at the phrase-level can improve constituency parsing, and (4) whether it is also effective on improving dependency parsing. Finally, we combine the priors to investigate the utility of concreteness at both the word and the phrase levels for further improving the performance of the \LPCFG{}. This combination allows more effect of the concreteness priors when training the \LPCFG{} to jointly learn the constituency and the dependency structures.
We experiment on the combined model that applies both priors either at the root level (as in \ours{}), or at the rule-level that produces the parsed tree $\bm{t}$ (as in \coupling{}) to fully understand the influence of the priors and the effect of where they are incorporated in the \LPCFG{} grammars.

\subsection{Dataset}
We use the MSCOCO dataset because it has a large number of visually concrete concepts, which provides a good testbed for our model. Although the MSCOCO dataset is not commonly used in previous works on dependency induction, MSCOCO has been used in previous work on visually informed constituency induction~\cite{shi2019visually, zhao-titov-2020-visually}. 
Using the same splits as VGNSL, the dataset contains 82,783 images for training, 1,000 for validation and testing, respectively. Each image has five corresponding captions. 
As a preprocessing step, we tokenize and lemmatize the captions (hyphenated words are preserved as a single token) using SpaCy \cite{spacy2}. We use the Berkeley constituent parser~\cite{Kitaev-2018-SelfAttentive} to create the gold trees, and we use the Stanford Parser for generating the gold dependencies~\cite{de-marneffe-manning-2008-stanford} for the validation and test sets. 


\paragraph{Concreteness scores} Each word of the processed sentence in the MSCOCO dataset is matched with its lemma's concreteness rating from \citet{concrete}. If a word's lemma is absent from the concreteness rating, the word has a concreteness of zero.


\subsection{Baseline Methods}
We compare our proposed \ours{} model and the \coupling~heuristic to several existing state-of-the-art methods:

\begin{itemize}[leftmargin=*]
\item\noindent \textbf{DMV}: The dependency model with valence \cite[DMV]{klein-manning-2004-corpus} is trained on POS tags obtained from an unsupervised tagging model following~\citet{DBLP:journals/corr/abs-1808-09111} using sentences that have fewer than 20 tokens. This allows the model to run faster while retaining high coverage of the data (97.6\% of the training set).

\item
\noindent \textbf{HI+FastText+IN}: The results from VGNSL with head-initial bias, FastText embeddings, and normalized image features~\cite{shi2019visually}.

\item
\noindent \textbf{1, S\textsubscript{MHI}, C\textsubscript{MX}-IN}: The results from \citet{kojima-etal-2020-learned} that reduces the image dimension to a single-dimension, uses mean and head-initial inductive bias for the score function (\textbf{S\textsubscript{MHI}}), uses max pooling for the combine function (\textbf{C\textsubscript{MX}}), and \textbf{-IN} removes normalized image features. 

\item 
\noindent \textbf{VC-PCFG}: The results of the visually-grounded compound PCFGs with the language modeling objective~\cite{zhao-titov-2020-visually} trained with 30 non-terminals, 60 preterminals, and 512-dimensional hidden states for the LSTM inference network.

\item
\noindent \textbf{\LPCFG{}}: The state-of-the-art neural L-PCFG~\cite{zhu-etal-2020-return}, i.e., the base model described in \autoref{sec:bg}.
\end{itemize}

\subsection{Our Methods}
To compare with previous work, we use pre-trained FastText \cite{arm2016fasttextzip} embeddings for the MSCOCO dataset on all variants of our model described in \autoref{sec:model}. 
We initialized the preterminals embeddings with centroids obtained using K-means clustering on the pre-trained word embeddings following \citet{zhu-etal-2020-return}.
This initialization allows the model to have a preliminary understanding of word meanings before training.

As described in \autoref{sec:model}, we experiment with three variants of our proposed model.
The first of these variants is \ours{}, described in \autoref{sec:concretemodel}. We experimented with different values of $\lambda_c$ in~\autoref{eq:main_eq} between $1.0$ and $3.0$, to vary the amplitude of the concreteness prior, but found that the model's performance is not strongly affected by this hyperparameter. 
Because the neural L-PCFG benefits from a larger grammar~\cite{zhu-etal-2020-return}, we increase the number of non-terminals and preterminals to 20 and 25, respectively, 
when training on the MSCOCO dataset as compared to the original paper. All other hyperparameters remain the same as~\citet{zhu-etal-2020-return}. 

We conduct experiments with the \coupling{} variant of the model, described in \autoref{sec:couplingmodel}. We train the model with 15 non-terminals, 20 preterminals, and decrease the hidden states of the LSTM inference network from 512 dimensions to 128. This is because, in our preliminary experiments, the effect of the vision-based heuristic was more apparent with a smaller inference network. We also run experiments with preterminals of 10 to further limit the strength of the neural LPCFG to uncover the difference, if any, in performance between dependency parsing and constituency parsing. One advantage of decreasing the grammar size is the reduction of parameters for faster training and lower memory requirements.
Two values, $1.0$ and $2.0$, are used for $\lambda_v$ in \autoref{eq:main_eq}.

Finally, we combine the two priors from \autoref{sec:concretemodel} and \autoref{sec:couplingmodel} in the \textsc{Combined\_Root} and \textsc{Combined\_Non\_Root} models, which perform the combination at the root-level and the rule-level, respectively. The combined models are trained with 15 non-terminals, 20 preterminals, and with the hidden states of 128 dimensions. Hyperparameters $\lambda_c$ and $\lambda_v$ are set to $0.25$ to avoid overpowering the \LPCFG{} with two sets of priors. 

We evaluate all models using directed and undirected attachment score (DAS and UAS) to measure dependency parsing, and F1 score for constituency parsing. We select the model checkpoint that maximizes the F1 score of the validation set. We report both corpus-level F1 and sentence-level F1 numbers in our experiments.

\begin{table}[tb]
\centering
\resizebox{\columnwidth}{!}{
\begin{small}
\huge
\begin{tabular}{llccccc@{}}
\toprule
    & PT    & NT  & \multicolumn{1}{c}{Corpus F1} & \multicolumn{1}{c}{Sent F1} &
\multicolumn{1}{c}{DAS} & \multicolumn{1}{c}{UAS} \\
\midrule
\textbf{Small grammar system} \\\midrule
\textsc{Neural L-pcfg} &  10 & 15  & {53.04} & 53.21 & 16.55 & 47.41\\ 
\coupling{} &  10 & 15 & \textbf{{54.84}} & \textbf{55.01} & {26.22} & \textbf{53.23} \\ 
\midrule
\textbf{Large grammar system} \\\midrule
\textsc{Neural L-pcfg} & 20 & 15 & {56.63} & {56.46} & {13.76} & {42.08} \\
\coupling{} &  20 & 15 & {58.52} & {58.6} & {18.27} & 45.79\\ 
\textsc{combined\_root} & 20 & 15  &	51.01 &	50.91 &	17.94 & 46.15 \\
\textsc{combined\_non\_root} & 20 &	15 & 	\textbf{68.73} & \textbf{69.23} &	18.22 &	{47.12} \\
\midrule 
\textsc{Neural L-pcfg}  & 25 & 20  & {58.02} & {58.63} & {15.63} & {42.33} \\
\ours{} & 25 & 20 & {54.81} & {55.21} & {\textbf{31.42}} & {\textbf{52.23}} \\ 
\midrule 
\textbf{Baselines}\\\midrule
DMV &  {-} & {-} & {-} & {-} & {18.79} & {42.47} \\
HI+FastText+IN & {-} & {-} & 54.4 & {-} & {-} &  {-} \\
1, S\textsubscript{MHI}, C\textsubscript{MX}-IN  & {-} & {-} & 57.5 & {-} & {-} & {-} \\
VC-PCFG** & 60 & 30 &  59.3 & 59.4 & {-} & {-} \\
\bottomrule
\end{tabular}
\end{small}
}
\caption{Performance of constituency parsing (F1) and dependency parsing (DAS/UAS) of our proposed models as compared to the baselines. The best setup of each model is reported in this table, selected based on validation set performance. The best scores for each task are \textbf{highlighted}. PT and NT are the number of preterminals and non-terminals of the \LPCFG, respectively. Priors are added at the root level for the \textsc{combined\_root} model. For the \textsc{non\_root} model, the priors are added at each rule-level. ** indicates the VC-PCFG~\cite{zhao-titov-2020-visually}, which is trained with a much larger grammar system. 
}
\label{tab:main_results}
\end{table}


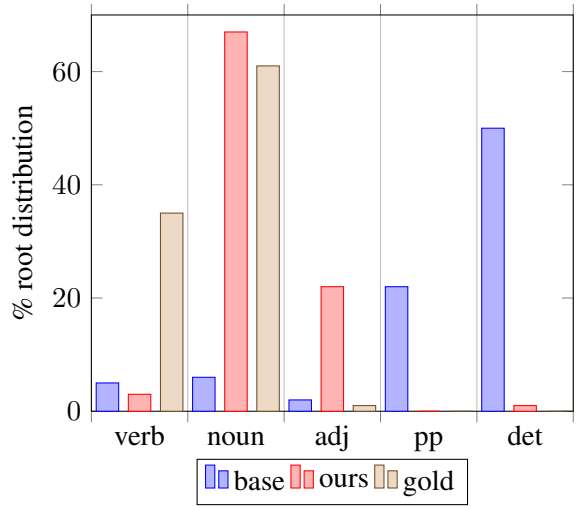
\begin{figure}[tb]
    \centering
\resizebox{\linewidth}{!}{
\begin{tikzpicture}
\begin{axis}[
	xtick={1,2,3,4,5,6},
	xticklabels={verb,noun,adj,pp, det, dummy},
	xticklabel style={text height=1ex},
	ylabel=\% root distribution,
	enlargelimits=0.00,
	legend style={at={(0.5,-0.12)},
	anchor=north,legend columns=-1},
	ybar interval=0.7,
	ymax=70
]
\addplot 
	coordinates {(1,5) (2,6)
		 (3,2) (4, 22) (5,50) (6,0)};
\addplot 
	coordinates {(1,3) (2,67) 
		(3,22) (4, 0) (5,01) (6,0)};
\addplot
	coordinates {(1,35) (2,61) 
		(3,1) (4, 0) (5,0) (6,0)};
		
\legend{base,ours,gold}
\end{axis}
\end{tikzpicture}
}
\caption{Root distribution (\%) by part-of-speech tags of the parse trees predicted with \LPCFG{} (base) and \ours{} (ours), compared to the gold distribution. POS categories with root distribution$<.05$for all setups are removed for clarity.
}
  \vspace{-2mm}
  \label{fig:pos}
\end{figure}

\begin{figure*}[t]
    \begin{subfigure}[t!]{0.3\textwidth}
    \centering
    \resizebox{\textwidth}{!}{%
        \begin{tikzpicture}
        \tikzset{edge from parent/.style=
                {draw,
                edge from parent path={(\tikzparentnode.south)
                -- +(0,-8pt)
                -| (\tikzchildnode)}}}
        \tikzset{level distance=60pt}
        \tikzset{every node/.style={font=\Huge}}
        \Tree [.\node(root){S-wildlife}; 
                [.NP-wildlife wildlife ]
                [.VP-standing standing [.PP-near near [.NP-area water area ] ]
                            [.PP-in in [.NP-setting natural setting ] ] ] ]
        \end{tikzpicture}%
    }
    \caption{Gold Tree}
    
    \end{subfigure}
    \begin{subfigure}[t!]{0.3\textwidth}
    \resizebox{\textwidth}{!}{%
    \begin{tikzpicture}
    \tikzset{edge from parent/.style=
            {draw,
            edge from parent path={(\tikzparentnode.south)
            -- +(0,-8pt)
            -| (\tikzchildnode)}}}
        \tikzset{level distance=60pt}
        \tikzset{every node/.style={font=\Huge}}
    \Tree [.\node(root){S-standing}; 
            [.NT-standing [.NT-standing wildlife standing ]
            [.NT-near near [.NT-area water area ] ] ]
            [.NT-in in [.NT-natural natural setting ] ] ]
    \end{tikzpicture} %
    }
    \caption{\textsc{Neural L-pcfg}}
    \label{sub:b}
    \end{subfigure}
    \begin{subfigure}[t!]{0.3\textwidth}
    \resizebox{\linewidth}{!}{%
    \begin{tikzpicture}
    \tikzset{edge from parent/.style=
            {draw,
            edge from parent path={(\tikzparentnode.south)
            -- +(0,-8pt)
            -| (\tikzchildnode)}}}
        \tikzset{level distance=60pt}
        \tikzset{every node/.style={font=\Huge}}
    \Tree [.\node(root){S-wildlife}; 
            [.NT-wildlife wildlife [.NT-standing standing [.NT-near near [.NT-area water area ] ] ] ]
            [.NT-in in [.NT-setting natural setting ] ] ] 
    \end{tikzpicture}  
    } 
    \caption{\textsc{combined}~(0.25)}
    \end{subfigure}
    \begin{subfigure}[b]{0.3\textwidth}
    \resizebox{\linewidth}{!}{%
    \begin{tikzpicture}
    \tikzset{edge from parent/.style=
            {draw,
            edge from parent path={(\tikzparentnode.south)
            -- +(0,-8pt)
            -| (\tikzchildnode)}}}
        \tikzset{level distance=60pt}
        \tikzset{every node/.style={font=\Huge}}
    \Tree [.\node(root){S-wildlife}; 
            [.NT-wildlife wildlife [.NT-standing standing [.NT-near near water ] ] area ]
            [.NT-in in [.NT-natural natural setting ] ] ] ]
    \end{tikzpicture}  
    }
    \caption{\coupling~(1.0)}
    \end{subfigure}
    \hfill
    \begin{subfigure}[b]{0.3\textwidth}
    \resizebox{\linewidth}{!}{%
    \begin{tikzpicture}
    \tikzset{edge from parent/.style=
            {draw,
            edge from parent path={(\tikzparentnode.south)
            -- +(0,-8pt)
            -| (\tikzchildnode)}}}
        \tikzset{level distance=60pt}
        \tikzset{every node/.style={font=\Huge}}
    \Tree [.\node(root){S-wildlife}; 
            [.NT-wildlife wildlife [.NT-standing standing [.NT-near near [.NT-area water area ] ] ] ]
            [.NT-in in [.NT-natural natural setting ] ] ]
    \end{tikzpicture}  
    }
    \caption{\coupling~(2.5)}
    \end{subfigure}
    \begin{subfigure}[b]{0.3\textwidth}
    \resizebox{\linewidth}{!}{%
    \begin{tikzpicture}
    \tikzset{edge from parent/.style=
            {draw,
            edge from parent path={(\tikzparentnode.south)
            -- +(0,-8pt)
            -| (\tikzchildnode)}}}
        \tikzset{level distance=60pt}
        \tikzset{every node/.style={font=\Huge}}
    \Tree [.\node(root){S-wildlife}; 
            [.NT-wildlife wildlife [.NT-standing standing [.NT-near near [.NT-water water area ] ] ] ]
            [.NT-in in [.NT-natural natural setting ] ] ]
    \end{tikzpicture}  
    }
    \caption{\textsc{Neural L-pcfg}**}
    \label{sub:f}
    \end{subfigure}
    \caption{The predicted tree structures for each model. ** indicates the tree structure was predicted by both \ours~and \LPCFG~with the same model size. Note that (b) is the baseline model in the vision-based heuristic experiments, and (f) is the baseline model in the concreteness experiments. 
    }
    \label{fig:bigtree}
\end{figure*}
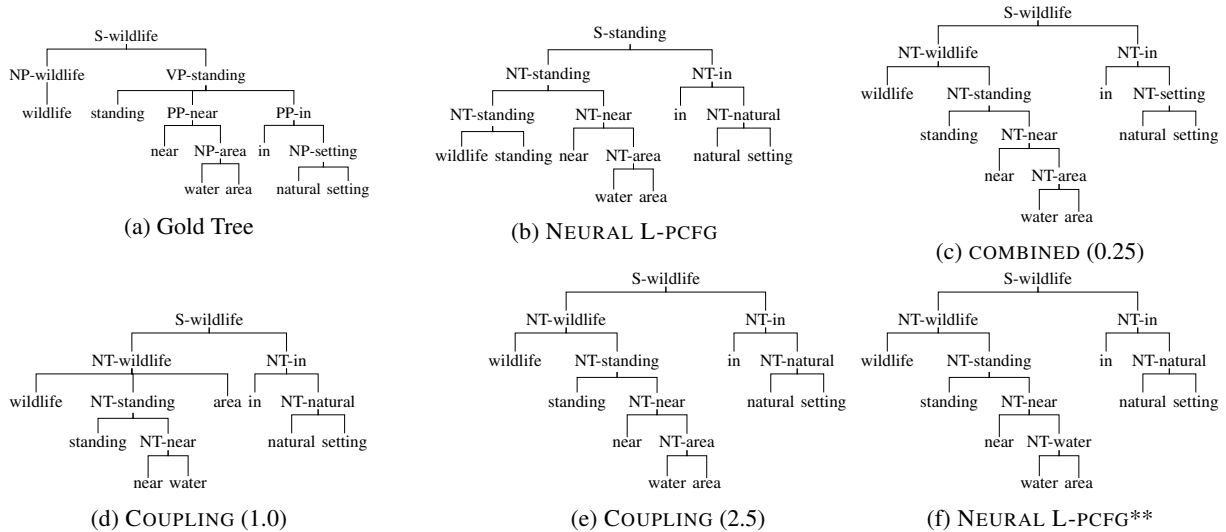

\subsection{Results}

\autoref{tab:main_results} shows our three model variants compared with the baseline models. Both the \ours{} and the \LPCFG{} show comparable results in F1 with the VGNSL from \citet{shi2019visually} and from~\citet{kojima-etal-2020-learned}. In dependency parsing, \ours{} doubles the DAS as compared to the \LPCFG{}, while also outperforming DMV by a large margin. 

For further analysis of the performance improvement, we look at the part-of-speech (POS) distributions of the sentence roots in MSCOCO (\autoref{fig:pos}). For the text-only \LPCFG, the model has a tendency to select prepositions and determiners as heads. This is intuitive because function words are far more common in the data, and there are no constraints against choosing them as the root of the tree. In contrast, \ours{} chooses nouns as the root word more often than function words, which is closer to the gold distribution, thus indicating the utility of the concreteness prior that upweights perceptible entities (typically nouns). 

In~\autoref{tab:main_results}, we observe that with smaller grammar systems, the proposed \coupling{} method achieves better dependency induction performance as compared to the \LPCFG{} of the same grammar size. 
With the large grammar system, the best \coupling{} model achieves comparable performance in constituency parsing (F1) with the \LPCFG{}, and all models improve performance in dependency parsing (DAS/UAS) from the baseline \LPCFG{} on MSCOCO. In addition, although the \textsc{Combined\_root} shows slightly lower performance in dependency parsing compared with the \textsc{Combined\_non\_root}, the \textsc{Combined\_root} still improves over the \LPCFG. This shows that enforcing the priors at the root level might be too late––the model has likely already selected incorrect heads in lower-level spans. 

Finally, we see that the \textsc{Combined} method outperforms all models in terms of constituency parsing performance by a large margin. We depict the predicted tree structures for an example sentence with various models in \autoref{fig:bigtree}. In this example, (b) \textit{standing} is incorrectly selected as the root of the sentence. Interestingly, (c) is almost identical to (e) except the correct head is predicted by (c) in the subtree \textit{natural setting}. It is possible that due to the simplicity of the constituency structures in MSCOCO, the neural L-PCFG finds a set of optimal latent rules which explains most of the noun-phrases in the corpus but fails to identify the correct root for dependency parsing. This indicates that learning the correct lexical head is rather difficult for the neural L-PCFG and it requires additional information such as the concreteness priors to improve on dependency parsing. 


\section{Discussion and Conclusion}
Overall, our results show that the substantial increase in dependency induction performance using our method is due to the added word concreteness prior. We demonstrate that concreteness is a good approximation for understanding subcategories of words in a dependency parsing model. However, we recognize that this approximation is insufficient for verb phrases, because an action described by a verb is often coupled with visible entities \cite{alikhani-stone-2019-caption}. For example, in a sentence \textit{many people gather around a building with clocks}, our approach incorrectly upweights \textit{people} to be the root rather than \textit{gather}, because the concreteness of the former is higher than the latter.

Similar to \citet{alikhani-stone-2019-caption}, our analysis in \autoref{fig:pos} shows that MSCOCO includes a large number of noun phrases (almost double the number of verb phrases). This disproportional ratio might contribute to the lower DAS for the DMV and text-only \LPCFG~models. Because MSCOCO contains shorter sentences with a simpler syntactic structure as compared to the Penn Treebank, \LPCFG~converges to a local optimum that achieves high constituency accuracy but low dependency accuracy (\autoref{tab:main_results}). By adding concreteness, our results show that dependency accuracy can be significantly improved while still maintaining a high constituency accuracy. 

Our experiments with the proposed vision-based heuristic show that both dependency and constituency accuracy measures can be improved by leveraging the joint learning setting of the \LPCFG{} and the visually derived information. Unlike previous work on visually grounded models, our \coupling{} heuristic selects what is visually significant from an image in the language modeling objective. However, certain limitations exist in our simple heuristic, including noise propagation from the semantic role labeling model. Because MSCOCO has a large portion of noun phrases, the action predicted by the semantic role labeling model from the image is often misaligned with a noun term in the caption and vice versa. This could mislead the downstream grammar induction task and hurt the constituency accuracy. For example, the left constituent from the root in (d) of \autoref{fig:bigtree} is grammatically incorrect. Given these observations, we removed the rewards to spans that couple the arguments when predicate is not aligned in the \textsc{Combined} model.

In future work, we plan to explore more elegant ways to parameterize the priors in the neural L-PCFG, instead of using the hyperparameter $\lambda$. Nonetheless, different from other visually grounded models, which jointly learn the textual representations with the paired image,
our approach directly influences the language model by encoding the visual information completely in the text domain. One advantage of our approach is that the visual information is directly applied to optimize a language modeling objective, allowing explainability of which parts of the image are helpful for learning the syntax of the corresponding caption. 

Our work provides a basis for solving higher-level syntax ambiguities. For example, the sentence \textit{the girl will put the orange on the tray in the bowl} can be decomposed in two ways: [the girl will put][the orange on the tray][in the bowl] and [the girl will put the orange][on the tray in the bowl]~\cite{doi:10.1080/17470218.2014.936475}. 
Although both analyses are grammatically correct, only one is correct when a visual reference (i.e., an image) is provided. Future dependency induction models could aim at untangling such ambiguity using more complex visual information.

Finally, although we show that concreteness is useful for dependency induction, using concreteness is seemingly an over-simplification for modeling semantic dependency relationships. In human language, root assignment can involve a priority mechanism: when a verb is present in a sentence, it immediately becomes the root despite how concrete other words are. Our work suggests that further investigation into modeling such priority mechanisms to improve the grammar induction performance even more.

\section*{Acknowledgements}
This work was supported in part by the DARPA GAILA project (award HR00111990063). The views and conclusions contained in this document are those of the authors and should not be interpreted as representing the official policies, either expressed or implied, of the U.S. Government. The U.S. Government is authorized to reproduce and distribute reprints for Government purposes notwithstanding any copyright notation here on. We would also like to thank the reviewers for their thoughtful comments.


\bibliographystyle{acl_natbib}
\bibliography{anthology,reference}


\end{document}